\documentclass[11pt, letterpaper]{article}
\usepackage[utf8]{inputenc}
\usepackage[bottom=2.0cm,top=2.0cm,left=2.0cm,right=2.0cm]{geometry}
\usepackage{graphicx}
\usepackage[utf8]{inputenc} % allow utf-8 input
\usepackage{fontenc}    % use 8-bit T1 font
\usepackage{times}
\usepackage{authblk}
\usepackage{amsmath}
\usepackage{float}
\usepackage{subfigure}
\usepackage{wrapfig}
\usepackage{svg}
\usepackage{hyperref}
\graphicspath{{figures/}}
 \def\hyph{-\penalty0\hskip0pt\relax}

\title{Heuristic Hyperparameter Optimization for Convolutional Neural Networks using Genetic Algorithm}
\author[1]{Meng Zhou}
\affil[1]{School of Computing, Queen's University, Kingston, ON, Canada}
\date{}

\begin{document}

    \maketitle
    \begin{abstract}

\noindent In recent years, people from all over the world are suffering from one of the most severe diseases in the history, known as Coronavirus disease 2019, COVID-19 for short. When the virus reaches the lungs, it has a higher probability to cause lung pneumonia and sepsis. X-ray is a powerful tool in identifying the typical features of the infection for COVID-19 patients. The radiologists and pathologists observe that ground-glass opacity appears in the chest X-ray for infected patient~\cite{cozzi2021ground}, and it could be used as one of the criteria during the diagnosis process. In the past few years, deep learning has proven to be one of the most powerful methods in the field of image classification. Due to significantly differences of Chest X-Ray between normal and infected people~\cite{rousan2020chest}, deep models could be used to identify the presence of the disease given a patient's Chest X-Ray. Many deep models are complex, and it evolves with lots of input parameters. Designers sometimes struggle with the tuning process for deep models, especially when they build up the model from scratch. Genetic Algorithm, inspired from biological evolution process, plays a key role in solving such complex problems. In this paper, I proposed a genetic-based approach to optimize the Convolutional Neural Network(CNN) for the Chest X-Ray classification task.

    \end{abstract}
    
    \section{Introduction}

The first massive outbreak of COVID-19 is in Wuhan, China, in December 2019, marked the beginning of crisis. Many people lose their life because of the severe disease and local economic is affected by the city lock down, no more public entertainments, even lots of the grocery stores, supermarkets are closed. The virus is very harmful to human body and it has a high infection rate. Some of the clinical symptoms of COVID-19 are muscle pain, cough, fever, difficulty breathing, etc. The standard COVID-19 test is very time consuming~\cite{mina2020rethinking}, and it is a drawback when facing dying patients. Hence, X-Ray could be used as a reliable alternative tool to test COVID-19 because it is cheaper and faster than the standard test. Nowadays, The diagnosis of COVID-19 involves the most common test called reverse transcription-polymerase chain reaction (RT-PCR), but sometimes we have to cross-validate for a particular result. Thus, medical imaging is the second choice for radiologists and doctors. Medical imaging, such as X-Ray and Computer Tomography (CT), plays a key role in detecting disease during early stage. Chest X-Ray has proven to be one of the effect methods to diagnose COVID-19 before the virus spread further to the lungs~\cite{shelke2021chest}. The suspicious COVID-19 cases could be observed from the Chest X-Ray images, hence it could also help to detect the rudimental disease.
\\
\\
Deep learning-based models have been proven to be efficient for the image classification task~\cite{he2020deep} and these models also have a wide application in medical image classification. For example, a recent grand challenge ``ProstateX"~\cite{litjens2017prostatex} shows the deep learning models have the ability to classify the presence of the Prostate Cancer by achieving the state-of-the-art performance. Deep models are trained to replace human in some redundant and regularly-happened tasks, while maintaining the same result quality. Thereby, deep models can be used as a computer aided diagnosis tool to perform specific tasks with high accuracy and speed. X-Ray images, unlike CT scan or standard COVID-19 test that are costly and hard to acquire in some rural areas, they are cheap and almost all clinical centers have X-Ray machines. With the vast amount of X-Ray images, it becomes feasible to train a deep model based on X-Ray images to help in detecting COVID-19 and could drastically reduce radiologists’ workloads.
\\
\\
Genetic based approaches~\cite{katoch2021review} have became a population method to optimize the best solution(s) given a complex problem. The method is inspired by the natural selection in biological systems, and follows the rule of ``survival of the fittest"~\cite{holland1992adaptation} during the evolution process. The new populations are generated by using the genetic operations on individuals that are presented in the original populations. Researchers have been studied Genetic Algorithms to solve mathematical optimization problems~\cite{hermawanto2013genetic}; to optimize the overall CNN architecture through neuroevolution strategy~\cite{stanley2002evolving} and to perform feature engineering as a preprocessing step prior training machine learning models~(\cite{pei1998feature}, \cite{adhao2020feature}). In this paper, I studied the application of genetic algorithm in automatically selecting the hyperparameters in the CNN. The task is a binary classification problem for determining whether a patient has COVID-19 or not based on the Chest X-Ray image. The detailed genetic algorithm design, network architecture and data could be found in Section~\ref{method}.

    \section{Related Work}

Medical Imaging has proven to be one of the effective tools for detecting diseases of a patient, such as prostate cancer, brain cancer, pneumonia, and the global pandemic COVID-19. Chest X-Ray could be use to detect COVID-19 by observing whether the ground-glass opacities and airspace opacities appear in the radiograph. Deep learning-based image classification tasks have achieved increasing research interests in the past few years and it is proven to be effect on large and complex data~\cite{khan2020survey}. Many researchers have focused their interests on the deep learning-based methods for developing Computer Aided Diagnosis(CAD) tools for detecting certain diseases using medical images~\cite{litjens2017survey}. A recent work from Shelke et al.~\cite{shelke2021chest} shows the deep learning models have the ability to diagnose and classify COVID-19 cases by achieving the state-of-the-art performance. They proposed a novel framework that combines VGG16, DenseNet161 and ResNet18 models to categorize the COVID-19 cases using Chest X-Ray data into three classes, namely severe, medium and mild. They used a total of 2271 high resolution X-ray images obtained from the Clinico Diagnostic Lab and then reduced to $64 \times 64$, gray-scale images for training and testing. The framework could achieve an accuracy score of 0.76 by using pre-trained weights on ImageNet dataset~\cite{deng2009imagenet} and fine tune for the COVID-19 classification task.
\\
\\
Genetic Algorithm(GA) based optimization has also been well studied to find the optimal solution from a huge search space~\cite{goldberg1988genetic}. Researchers have attempted to apply GA to CNN models to optimize the model architecture or perform features selection after learnt by deep models. In a recent study from Yilmaz et al.~\cite{yilmaz2020novel}, they used a hybrid framework to recognize human actions across four different datasets: Skoda, UCI Smartphone, KTH, UCF Sports Action. The GA method is utilized as an optimization step to select importance features prior passing into Fully Connected(FC) layers for classification. In detail, they used four different backbones: AlexNet, VGG16, Google Net and ResNet152 to extract the features for a given image. Each backbone will output 2048 features in the latent space that represents the input image, resulted in 8192 features in total. Then, the GA method is designed to reduce the total features to 4102 features by selecting best features through optimization. The archiecture of GA follows the classical setting, composed with initialization, fitness evaluation, parent selection, mutation, crossover and survivor selection. Finally, 4102 features are feeding into FC layers to classify the input human action to 66 categories. With GA as an optimization step, the proposed model could achieve the state-of-the-art performance on human actions recognition across four datasets.

    \section{Method} \label{method}

\begin{figure*}[t]
	\begin{center}
		%\fbox{\rule{0pt}{2in} \rule{.9\linewidth}{0pt}}
		\includegraphics[width=.99\textwidth]{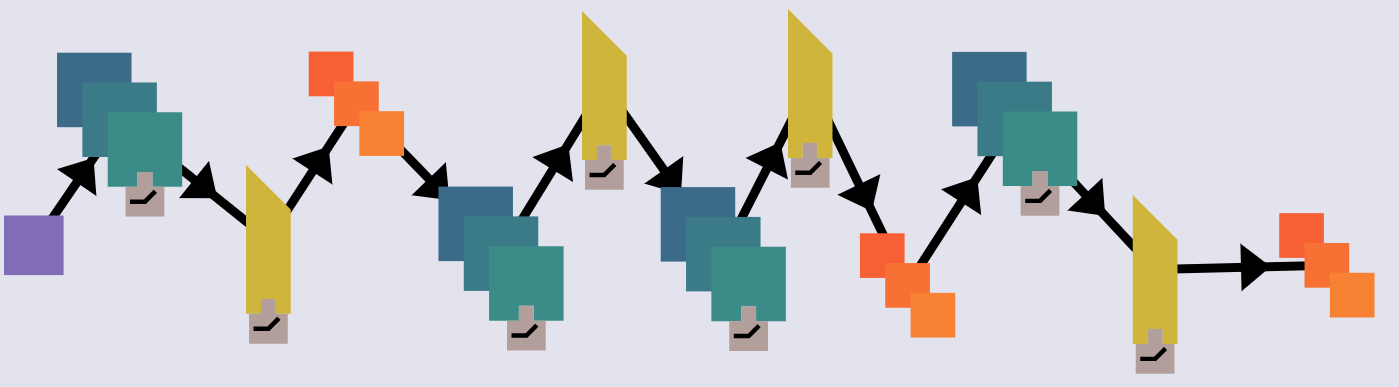}
	\end{center}
	\vspace{-3ex}
	\caption{Feature Extractor of the proposed CNN architecture}
	\label{455cnn}
	\vspace{-3ex}
\end{figure*}

My goal is to find a set of optimal parameters for the pre-defined CNN architecture based on Genetic Algorithm. Due to the limited of computational resources and the small amount of data described in Section~\ref{data}, the CNN architecture only has four Convolutional layers, two FC layers with random dropout and one Softmax layer with output dimension of 2, indicating the binary classification task. Figure~\ref{455cnn} only shows the feature extractor of the proposed architecture. The purple square in the left denotes the input batch of images, following by a Convolutional layer denoted in 
aquamarine, the orange square is the Max Pooling layer and the yellow trapezoid denotes the Batch Normalization layer. After the right-most Max Pooling layer, a Adaptive Average Pooling layer is applied for the consistency output from the feature extractor. The output features from the feature extractor pass into FC layers to produce the probabilities on how likely the image belongs into the two categories. The output channels, kernel size of Convolutional layers; all activation functions; features in FC layers and drop out rate in Dropout layers are selected by the GA. More details of GA implementations are in Section~\ref{ga}.

    \subsection{Data} \label{data}

The data \footnote{\url{https://www.kaggle.com/c/cisc873-dm-f20-a2/data}} used in this study is small, and is collected from different sources. There are total of 487 chest X-Ray images with labeled either 0 or 1 for training, around 300 of them are positive {\it i.e.} detected for COVID-19, and rest of them are negative {\it i.e.} normal lung X-rays. Training images are resized to $256 \times 256 \times 3$, normalized the pixel values to $[0,1]$ for all images, no data augmentation technique performed due to limited computational resources and the costly of GA optimization time. The images are then splitted to training and validation data with ratio $0.8:0.2$, and the validation accuracy is used as the selection criteria for picking the best model.

    \subsection{GA Design} \label{ga}

\begin{figure*}[t]
	\begin{center}
		%\fbox{\rule{0pt}{2in} \rule{.9\linewidth}{0pt}}
		\includegraphics[width=.99\textwidth]{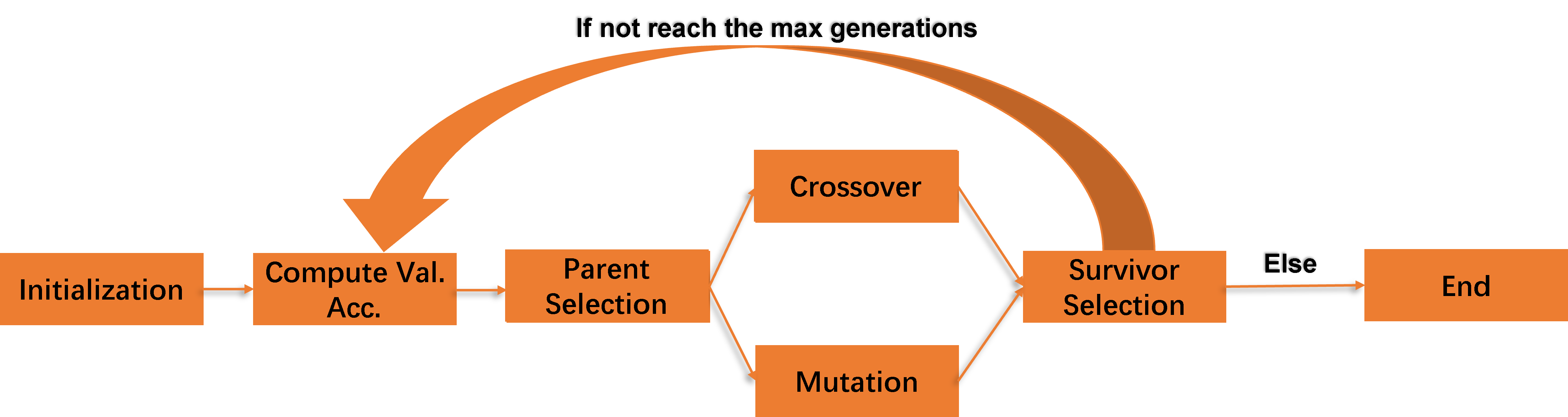}
	\end{center}
	\vspace{-3ex}
	\caption{Genetic Algorithm Flowchart}
	\label{455gf}
	\vspace{-3ex}
\end{figure*}

\begin{figure*}[t]
	\begin{center}
		%\fbox{\rule{0pt}{2in} \rule{.9\linewidth}{0pt}}
		\includegraphics[width=.99\textwidth]{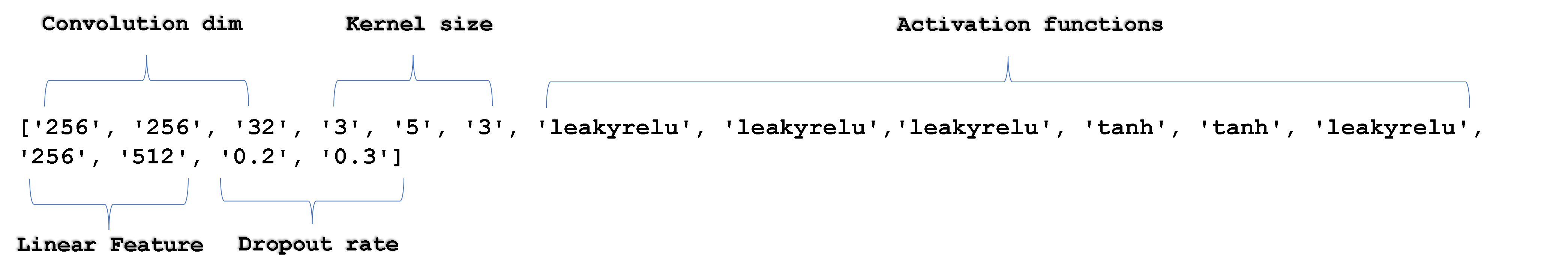}
	\end{center}
	\vspace{-3ex}
	\caption{Parameters interval within a single parent}
	\label{455cross}
	\vspace{-3ex}
\end{figure*}

In this section, the detailed GA implementation is discussed, the algorithm consists of initialization, fitness evaluation, mutation, crossover, survivor and parent selection. The flowchart of the GA is shown in Figure~\ref{455gf} .
\\
\\
\textbf{Initialization.} The initialization step of GA is to generate the CNN required parameters randomly. Recall that in Figure~\ref{455cnn}, the pre-defined CNN model has four Convolutional layers. For simplicity, I fix the first Convolutional layer to have the input channels of 3 and output channels of 32 with a $3 \times 3$ kernel filter. The remaining output channels are randomly selected from $[32, 64, 128, 256]$, and kernel sizes are randomly selected from $[3,5,7]$. For FC layers in the final classifier, the linear features in each FC layer are also randomly selected from $[512, 256, 128]$. Activation functions, for all layers, are randomly selected from $[tanh, relu, leakyRelu]$, which are the three most common nonlinear activation functions in neural networks. Each FC layer may associates with an additional Dropout layer, the dropout rates are randomly selected from $[0.1, 0.2, 0.3, 0.5]$. To conclude, there are total of 16 (3 convolution dim + 3 kernel + 6 activation functions + 2 linear features + 2 dropout) parameters to be selected for each parent. The initialization function returns a two\hyph{dimensional} array with size $N \times 16$ where $N$ is the number of initial population and a index counter to store the index range of each parameter. Each parent is mapped to a specific CNN model under this setting.
\\
\\
\textbf{Fitness Evaluation.} The fitness is computed as the validation accuracy of the individual from the initial population. Each model is generated from the parent in the pool with the corresponding parameters. The training data loader has the size of $[390, 3, 256, 256]$ where 390 is the total number of training images; similarly, the validation data loader has the size of $[97, 3, 256, 256]$ where 97 is the total number of validation images. All models are trained for 20 epochs with constant learning rate of 0.0005 to prevent overfitting, the Stochastic Gradient Descent with Adam~\cite{kingma2014adam} update rule is used as the optimizer. The batch size of the data loader is 16 due to limited memory available. The loss function is the weighted Categorical Cross Entropy due to the extreme imbalance data samples between two classes. The loss function and the batch size remains the same across all models. The fitness evaluation function returns a list of validation accuracy for each parent in the initial population.
\\
\\
\textbf{Parent Selection.} The tournament selection is used for choosing the best parents. In each tournament, five parents are randomly selected to compete with each other and the winner is picked for to produce offspring and perform crossover. After the winner is picked, it will be removed from the current population before next tournament begins. If multiple parents have the same fitness score, then randomly breaks the tie. In each generation, ten parents from ten independent tournaments are selected. The function returns a two\hyph{dimensional} list containing selected parents.
\\
\\
\textbf{Crossover.} A level\hyph{wise} crossover method is utilized in this study. In each parent, the interval of each parameter is shown in Figure~\ref{455cross}. The idea of crossover is to swap the corresponding interval between parents. This encourages the population pool to maintain the diversity as two parents will switch these parts. If the convolution dimension or linear feature part is selected, then the target model will have different feature extractor or classifier than before (assume these parts are different initially). Two\hyph{point} crossover is performed with crossover rate of 0.6, two distinct parts of the individual are selected randomly, and swap between parents in a sequential manner. The crossover function returns two new parents, each of them is a one\hyph{dimensional} list. 
\\
\\
\textbf{Mutation.} The multi\hyph{point} mutation is performed. The first mutation is to tackle with the Convolutional layer input and output dimension. Since the input dimension of a Convolutional layer is the output dimension of the previous layer, the consideration of the output dimension is enough. To speed up the GA optimization process and let the model converges with in 20 epochs, I used a greedy approach to rearrange the dimensions. A classical and efficient way is to let the convolution dimensions to increase as the number of layers increase {\it i.e.} the network learns more low\hyph{level} features as the number of layers increase. Hence, in the first mutation round, the dimension for three Convolutional layer has been mutated in the increasing order. In the actual mutation step, one of the convolution dimension, kernel size, activation functions, linear features and dropout rate is selected randomly, then the mutation is performed within that part. The algorithm performed mutation in every generation. The mutation functions returns a one\hyph{dimensional} list representing the mutated offspring.
\\
\\
\textbf{Survivor Selection.} The basic $\lambda + \mu$ selection method is performed. The worst $\lambda$ parents {\it i.e.} low validation accuracy are replaced by the $\lambda$ new generated offspring. This function returns the new population for the next generation.

    \section{Experiments}

The GA is implemented in Python 3.6.13, and the CNN is implemented through the same Python version and PyTorch 1.2.0. The GA stops when it reaches to 100 generations, and it has has been executed for several times independently for different population size. The highest validation accuracy of each population size is reported as the final result. The average run time for one complete GA is about 10 hours. The class weights are computed as in the Equation~\ref{cw} , where $N$ is the total number of samples and $m_{i}$ is the number of samples in class $i$. The additional factor of 2 in the denominator restrains the loss to a similar magnitude with the loss generated from no class weights. the best validation accuracy achieved is 0.749. Notice that the validation accuracy increases as the initial population size increases, showing that preserving diversity in the initial population is crucial. For fair comparison, a random seed is set when splitting training and validation data so that in each independent run, all models will have the same training image.

\begin{equation}
\begin{split}{
    W_{i} = \frac{N}{2 \times m_{i}} \\
}\end{split}
\label{cw}
\end{equation}
    
    \subsection{Comparison}

As the baseline result, I used the pre-trained VGG16~\cite{simonyan2014very} model and fine tuned the last FC layer in the classifier for this task. The base feature extractor used the pre-trained weights on the ImageNet dataset~\cite{deng2009imagenet} and the weights are freezed while fine tuning. The final FC layer of VGG16~\cite{simonyan2014very} is changed to 4096 features as input and 2 features as output. The learning rate is 0.0005, batch size is 16, and the model is trained for 20 epochs with Adam~\cite{kingma2014adam} optimizer. The validation accuracy achieved by the pre\hyph{trained} VGG16 model is 0.688. The result shows that the model generated by the GA has overperformed the pre\hyph{trained} model, indicating the effectiveness of the GA in this few\hyph{shot} classification task.

\begin{table}[H]
%\caption{my table}
\centering
\begin{tabular}{|c|c|}
\hline
%\multicolumn{1}{|c|}{\multirow{1}*{\diagbox{}{}}} &\multicolumn{1}{|c|}{Method} &\multicolumn{1}{|c|}{T=10000} &\multicolumn{1}{|c|}{T=100,000}\\
Method& Val. Acc.\\ 
\hline
%\multicolumn{2}{|c|}{} & (2,3) & (2,4) & (2,5) & \\
%\hline
GA, $N_{p}$=50 & 0.749\\
GA, $N_{p}$=30 & 0.664\\
GA, $N_{p}$=20 & 0.656\\
\cline{1-2}
Pre-trained VGG16 & 0.688\\
\hline
\end{tabular}
\caption{ Quantitative results for two methods, $N_{p}$ is the initial population size of the GA }
\end{table}

    \section{Limitations and Discussions}

In this section, the advantages and drawbacks of GA, the dataset issue and the possible future work is discussed.

    \subsection{Analysis of GA}
Through this study, the GA-based optimization method for selecting CNN parameters is proven to be successful in the classification of the COVID-19. The GA with initial population size 50 achieves the best validation accuracy score. This also shows CNN works excellent with the image data. Its unique two-dimensional filters were used to catch the local features of each image to be able to distinguish between the positive and negative COVID-19 cases. The optimal set of parameters selected by GA is reliable based on the result and it follows the structure of common modern CNN parameters: the output dimensions of Convolutional layers increase as the number of layers increase; the activation functions of Convolutional layers are belong to either $ReLU$ or $LeakyReLU$; the linear features decreases as the number of FC layers and the drop rate for the last FC layers is 0.3. One thing to mentioned is the output dimensions for Convolutional layers are not always be the number of power of 2, which is different from the common choice of output dimensions {\it i.e.} $64, 256, 512$, etc. In the optimal set of parameters, the output dimensions for three Convolutional layers are 81, 116 and 160, respectively. This result shows that an alternative choice of output dimensions in Convolutional layers. The same situation applies to the linear features in FC layers. In terms of the efficiency of GA, the algorithm tends to find the local optimal solution in a short amount of time, and costs much more time when it tries to step out from the current local optimal and search for the another local or global optimal solution. 

    \subsection{Data Issue}

It is risky to pull out around 100 images as validation data when train the model, because the data is limited and thus less features will be learnt by the model. However, to make this task as much as close to the real\hyph{life} scenario, the validation data is necessary to evaluating the model. As a result, the model performance is affected by this issue, regardless the set of parameters generated by GA. To address the data limitation problem, data augmentation is a feasible way to boost the number of training data and prevent overfitting, a common method is to apply random affine transformations to the data {\it i.e.} rotate, flip, crop, etc.. However, the GA optimization time for each generation will be extremely long if augmentation is applied, making this method infeasible to achieve under this situation.

    \subsection{GA Drawbacks}

One major disadvantage of the proposed GA in this study is the CNN architecture must be pre-defined, this means the structure of CNN models could not be changed when GA starts to search the best solution. If a different task is assigned, the current CNN may not be used again. Then the GA has to be changed, in other words, the GA is task\hyph{dependent} and hence it has poor generalizability. Another point is the mutation operation I mentioned in Section~\ref{ga}, I used a greedy approach to changed the order of output dimensions of Convolution layers in a increasing order. This is because I want to have a promising result within a limited number of generations. Normally, with sufficient computational resources, one could mutate one layer at a time and set the maximum generation to a larger number than 100.

    \subsection{Future Work}

Based on the issues and drawbacks addressed above, there are few things could work on in the future. First, one could consider data augmentation techniques to improve the model performance. Alternatively, there are other Chest X-Ray data for COVID-19 diagnosis, one could look for other related data and combine them together to form a larger dataset. In either way, the GA-based optimization process will take much longer time, since the amount of data increases, and the training and validating time also increases. For the CNN architecture, one could modify it based on different tasks, or could alternatively use the method proposed in Stanley et al.~\cite{stanley2002evolving} to automatically design the CNN architecture by the GA. In the meantime, the GA functions, such as initialization, fitness evaluation, crossover, etc. should be modified accordingly. Last but not least, the GA parameters, such as crossover rate, initial population size and maximum generations, could be further adjusted. The current setting is intuitive, and may not be the optimal one for all tasks. Nevertheless, the proposed GA approach is simple, effective, and I believe my method can be applied to different kinds of image classification tasks with limited amount of data.

    \section{Acknowledgements}

This study is the final project of CISC455, Evolutionary Optimization and Learning course at Queen's University, Canada. The code is available at~\url{https://github.com/simonZhou86/EvolutionaryComputing}. The code is not designed for, and should not be used for any other academic purposes in any circumstances. However, the extension of this work is encouraged.

\bibliographystyle{unsrt} % Style BST file
\bibliography{cisc455}

\begin{thebibliography}{10}

\bibitem{cozzi2021ground}
Diletta Cozzi, Edoardo Cavigli, Chiara Moroni, Olga Smorchkova, Giulia
  Zantonelli, Silvia Pradella, and Vittorio Miele.
\newblock Ground-glass opacity (ggo): a review of the differential diagnosis in
  the era of covid-19.
\newblock {\em Japanese journal of radiology}, pages 1--12, 2021.

\bibitem{rousan2020chest}
Liqa~A Rousan, Eyhab Elobeid, Musaab Karrar, and Yousef Khader.
\newblock Chest x-ray findings and temporal lung changes in patients with
  covid-19 pneumonia.
\newblock {\em BMC Pulmonary Medicine}, 20(1):1--9, 2020.

\bibitem{mina2020rethinking}
Michael~J Mina, Roy Parker, and Daniel~B Larremore.
\newblock Rethinking covid-19 test sensitivity—a strategy for containment.
\newblock {\em New England Journal of Medicine}, 383(22):e120, 2020.

\bibitem{shelke2021chest}
Ankita Shelke, Madhura Inamdar, Vruddhi Shah, Amanshu Tiwari, Aafiya Hussain,
  Talha Chafekar, and Ninad Mehendale.
\newblock Chest x-ray classification using deep learning for automated covid-19
  screening.
\newblock {\em SN computer science}, 2(4):1--9, 2021.

\bibitem{he2020deep}
Zhengyu He.
\newblock Deep learning in image classification: A survey report.
\newblock In {\em 2020 2nd International Conference on Information Technology
  and Computer Application (ITCA)}, pages 174--177. IEEE, 2020.

\bibitem{litjens2017prostatex}
Geert Litjens, Oscar Debats, Jelle Barentsz, Nico Karssemeijer, and Henkjan
  Huisman.
\newblock Prostatex challenge data.
\newblock {\em The cancer imaging archive}, 10:K9TCIA, 2017.

\bibitem{katoch2021review}
Sourabh Katoch, Sumit~Singh Chauhan, and Vijay Kumar.
\newblock A review on genetic algorithm: past, present, and future.
\newblock {\em Multimedia Tools and Applications}, 80(5):8091--8126, 2021.

\bibitem{holland1992adaptation}
John~Henry Holland et~al.
\newblock {\em Adaptation in natural and artificial systems: an introductory
  analysis with applications to biology, control, and artificial intelligence}.
\newblock MIT press, 1992.

\bibitem{hermawanto2013genetic}
Denny Hermawanto.
\newblock Genetic algorithm for solving simple mathematical equality problem.
\newblock {\em arXiv preprint arXiv:1308.4675}, 2013.

\bibitem{stanley2002evolving}
Kenneth~O Stanley and Risto Miikkulainen.
\newblock Evolving neural networks through augmenting topologies.
\newblock {\em Evolutionary computation}, 10(2):99--127, 2002.

\bibitem{pei1998feature}
M~Pei, ED~Goodman, and WF~Punch.
\newblock Feature extraction using genetic algorithms.
\newblock In {\em Proceedings of the 1st International Symposium on Intelligent
  Data Engineering and Learning, IDEAL}, volume~98, pages 371--384, 1998.

\bibitem{adhao2020feature}
Rahul Adhao and Vinod Pachghare.
\newblock Feature selection using principal component analysis and genetic
  algorithm.
\newblock {\em Journal of Discrete Mathematical Sciences and Cryptography},
  23(2):595--602, 2020.

\bibitem{khan2020survey}
Asifullah Khan, Anabia Sohail, Umme Zahoora, and Aqsa~Saeed Qureshi.
\newblock A survey of the recent architectures of deep convolutional neural
  networks.
\newblock {\em Artificial Intelligence Review}, 53(8):5455--5516, 2020.

\bibitem{litjens2017survey}
Geert Litjens, Thijs Kooi, Babak~Ehteshami Bejnordi, Arnaud Arindra~Adiyoso
  Setio, Francesco Ciompi, Mohsen Ghafoorian, Jeroen~Awm Van Der~Laak, Bram
  Van~Ginneken, and Clara~I S{\'a}nchez.
\newblock A survey on deep learning in medical image analysis.
\newblock {\em Medical image analysis}, 42:60--88, 2017.

\bibitem{deng2009imagenet}
Jia Deng, Wei Dong, Richard Socher, Li-Jia Li, Kai Li, and Li~Fei-Fei.
\newblock Imagenet: A large-scale hierarchical image database.
\newblock In {\em 2009 IEEE conference on computer vision and pattern
  recognition}, pages 248--255. Ieee, 2009.

\bibitem{goldberg1988genetic}
David~E Goldberg and John~Henry Holland.
\newblock Genetic algorithms and machine learning.
\newblock 1988.

\bibitem{yilmaz2020novel}
Abdullah~Asim Yilmaz, Mehmet~Serdar Guzel, Erkan Bostanci, and Iman Askerzade.
\newblock A novel action recognition framework based on deep-learning and
  genetic algorithms.
\newblock {\em IEEE Access}, 8:100631--100644, 2020.

\bibitem{kingma2014adam}
Diederik~P Kingma and Jimmy Ba.
\newblock Adam: A method for stochastic optimization.
\newblock {\em arXiv preprint arXiv:1412.6980}, 2014.

\bibitem{simonyan2014very}
Karen Simonyan and Andrew Zisserman.
\newblock Very deep convolutional networks for large-scale image recognition.
\newblock {\em arXiv preprint arXiv:1409.1556}, 2014.

\end{thebibliography}
\end{document}